\documentclass[letterpaper, 10 pt, conference]{ieeeconf}
\IEEEoverridecommandlockouts % This command is to use the \thanks command
\usepackage[T1]{fontenc}
\usepackage[cmex10]{amsmath}
\usepackage{amssymb}
\usepackage{graphicx}
\usepackage{subfigure}
\usepackage{balance}
\usepackage{lipsum}
\usepackage[ruled,vlined]{algorithm2e}
\usepackage{url}
\usepackage{cite}
\usepackage{xcolor}
\usepackage{xfrac}
\usepackage{soul}
\let\labelindent\relax
\usepackage{enumitem}

\newcommand\figurename{Fig.}
\newcommand\set[1]{\mathcal{#1}}

\DeclareMathOperator*{\argmax}{\arg\max}

\newtheorem{defn}{Definition}

\begin{document}

%% Paper meta-information
\title{\LARGE \bf Learning Human-Robot Handovers Through
  $\mathbf{\pi}$-STAM:\\ Policy Improvement With Spatio-Temporal
  Affordance Maps}

\author{Francesco Riccio$^1$* \and Roberto Capobianco$^1$* \and
  Daniele Nardi$^1$%
  \thanks{$^1$Francesco Riccio, Roberto Capobianco and Daniele Nardi
    are with the Department of Computer, Control, and Management
    Engineering, Sapienza University of Rome, via Ariosto 25, Rome,
    00185, Italy {\tt\small \{riccio, capobianco, nardi\}@dis.uniroma1.it}}%
  \thanks{* These two authors equally contributed to the work.}%
} %

%978-1-4673-9163-4/15/$31.00 ©2015 IEEE
% \IEEEpubid{\makebox[\columnwidth]{\hfill\bf
%     978-1-4673-9163-4/15/\$31.00~\copyright~2015
%     IEEE}\hspace{\columnsep}\makebox[\columnwidth]{}}

\maketitle

%%%%%%%%%%%%%%%%%%%%%%%%%%%%%%%%%%%%%%%%%%%%%%%%%%%%%%%%%%%%%%%%%%%%%%%%%%%%%%%%

\begin{abstract}
  Human-robot handovers are characterized by high uncertainty and poor
  structure of the problem that make them difficult tasks. While
  machine learning methods have shown promising results, their
  application to problems with large state dimensionality, such as in
  the case of humanoid robots, is still limited. Additionally, by
  using these methods and during the interaction with the human
  operator, no guarantees can be obtained on the correct
  interpretation of spatial constraints (e.g., from social rules). In
  this paper, we present Policy Improvement with Spatio-Temporal
  Affordance Maps -- $\pi$-STAM, a novel iterative algorithm to learn
  spatial affordances and generate robot behaviors. Our goal consists
  in generating a policy that adapts to the unknown action semantics
  by using affordances. In this way, while learning to perform a
  human-robot handover task, we can (1) efficiently generate good
  policies with few training episodes, and (2) easily encode action
  semantics and, if available, enforce prior knowledge in it. We
  experimentally validate our approach both in simulation and on a
  real NAO robot whose task consists in taking an object from the
  hands of a human. The obtained results show that our algorithm
  obtains a good policy while reducing the computational load and time
  duration of the learning process.
\end{abstract}

%%%%%%%%%%%%%%%%%%%%%%%%%%%%%%%%%%%%%%%%%%%%%%%%%%%%%%%%%%%%%%%%%%%%%%%%%%%%%%%%

\section{Introduction}
\label{sec:intro}
Human-robot handovers are fundamental but difficult tasks, due to the
unpredictability of human behaviors and the need to appropriately
recover from robot failures. They are, in fact, characterized by high
uncertainty and poor structure of the problem that make it difficult
and time consuming to hand-craft appropriate policies, especially in
the case of complex humanoid robots. To tackle this problem, machine
learning -- and in particular, reinforcement learning -- has been
increasingly used
\cite{Wilson2012,Kalakrishnan2013,Gritsenko2014}. Nevertheless, to
learn from data a (sub-)optimal set of parameters for the generation
of robot behaviors, a huge number of learning episodes is required due
to the curse of dimensionality caused by a typically very large
state-space. Additionally, while the human operator needs to be
involved or at least simulated during the learning process, guarantees
on the correct interpretation of spatial constraints can be obtained
only at the price of modeling a more complex
problem~\cite{Holladay2016}. For example, in the case of humanoid
robots, such constraints may arise from social rules, which must be
respected to make the robots acceptable in human populated
environments.

During the learning phase of human-robot handovers, and more in
general of robot policies, environment-based interpretations of
actions can be highly beneficial to reduce the search space and
eventually enforce constraints on the robot behavior. However, rather
than spaces or more generally environments, previous literature only
considers action semantics in relation to objects. For example, the
concept of affordances as action opportunities that objects
offer~\cite{Gibson1979} has been used in robotics to
learn~\cite{Koppula2013}, represent~\cite{Pandey2012} and
exploit~\cite{Kim2015} object related actions in human-populated
environments. While in these works affordances do not change over time
and they are statically attached to each object, the state of the
environment is highly dynamic and also contains several dynamic
entities, such as humans and robots themselves. This inevitably leads
to a more complex problem that requires specific representation and
learning approaches, such as the Spatio Temporal Affordance Maps
(STAM) proposed in~\cite{Riccio2016}. By using STAM, action
affordances change according to the state of the environment and the
task to be executed. For example, while executing handovers with
humans, the robot may need to wait according to the attention of the
partner person -- e.g., wait for eye contact.

\begin{figure}[!t]
  \centering
  \includegraphics[width=0.8\columnwidth]{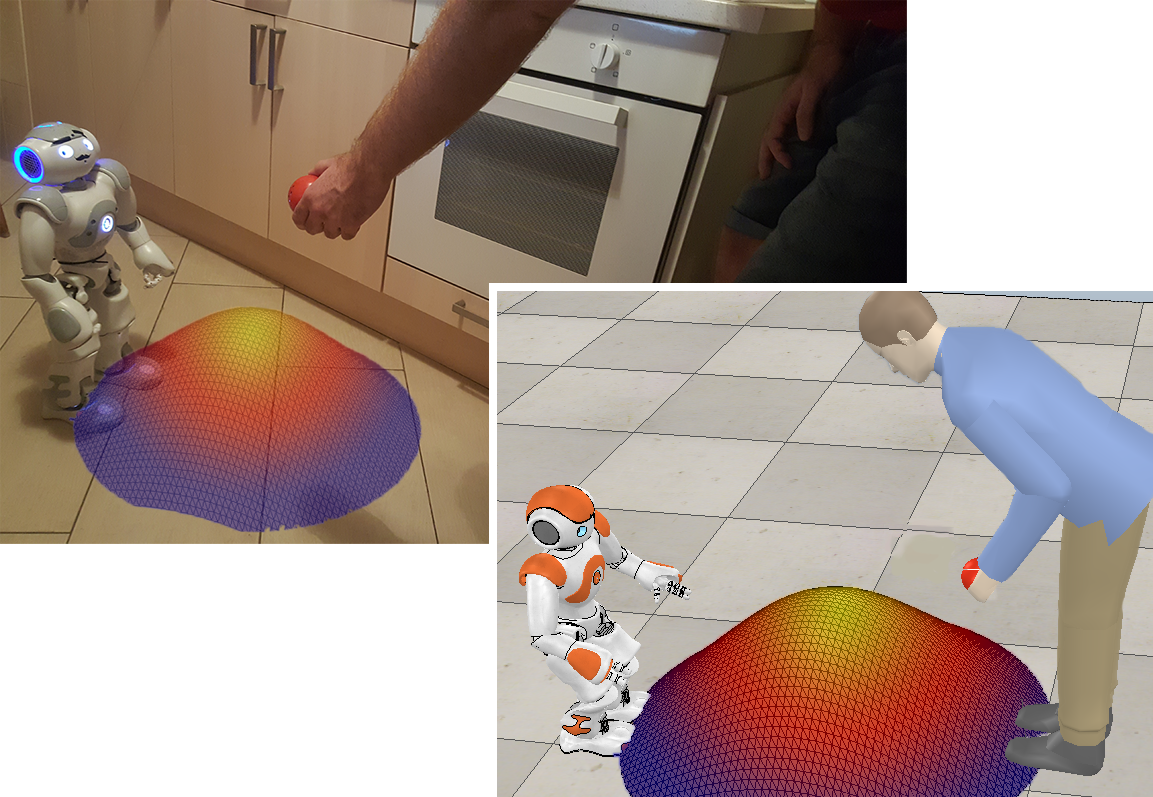}
  \caption{Learned affordance function for the action
    \texttt{left\_arm\_up} both in a simulated environment and real
    scenario. In the current state the action is not legal, since the
    likelihood to afford \texttt{left\_arm\_up} is low. Conversely, the
    action becomes legal when the robot distance from the target object
    decreases.}
  \label{fig:intro}
\end{figure}

In this paper, we present Policy Improvement with Spatio-Temporal
Affordance Maps ($\pi$-STAM), a novel iterative algorithm to learn
spatial affordances and generate robot behaviors. Our goal consists in
generating a policy that is composed of a discrete set of actions and
adapts to the unknown action semantics of the environment. As in
MCSDA~\cite{Riccio2016a}, our algorithm uses a standard classifier to
generate a policy, that is initialized to be random and it is
continuously refined by aggregating~\cite{Ross2011} the initial
dataset with rollouts collected through Monte Carlo
search~\cite{Browne2012}. $\pi$-STAM is a model-based reinforcement
learning approach that, however, builds on methods with unknown system
dynamics~\cite{Ross2014} and additionally uses the aggregated dataset
to improve the spatio-temporal affordance maps of the environment. As
shown in \figurename~\ref{fig:intro}, such maps can then be used to
generate a set of legal actions, hence reducing the search space of
the Monte Carlo algorithm and modifying robot behaviors based on
high-level information. With this paper, in fact, we aim at showing
that by using affordances to learn a human-robot handover task (1) we
can reduce the learning time and efficiently generate good policies
with few training episodes, and (2) we can easily encode action
semantics in the operational scenario and, if available, enforce prior
knowledge in robot behaviors. We experimentally validate our approach
both on simulated and real NAO robots, whose task consists in taking
an object from the hands of a human operator. The obtained results
show the effectiveness of $\pi$-STAM in obtaining a good policy while
reducing the computational load and time duration of the learning
process.

The reminder of this paper is organized as
follows. Section~\ref{sec:related_work} provides an overview of the
literature about affordances and policy learning;
Section~\ref{sec:stam} describes the concept of Spatio-Temporal
Affordance Maps, and Section~\ref{sec:pi_stam} presents in detail the
proposed approach, introducing the $\pi$-STAM algorithm. Finally,
Section~\ref{sec:evaluation} describes the experimental setup and the
results obtained both in simulation and on the real robot;
Section~\ref{sec:discussion} concludes the paper with some final
remarks and future work.

%%%%%%%%%%%%%%%%%%%%%%%%%%%%%%%%%%%%%%%%%%%%%%%%%%%%%%%%%%%%%%%%%%%%%%%%%%%%%%%%

\section{Related Work}
\label{sec:related_work}

In this paper, our goal is to generate a policy for the execution of
human-robot handovers by using a NAO humanoid platform. Manifold
machine learning have been adopted to achieve collaboration between
humans and robots.
%For example, Wang et al.~\cite{Wang2012} allow a
%robot to react to human actions by relying upon probabilistic Bayesian
%classification of human intentions. Equivalently, Ashesh et
%al.~\cite{Ashesh2013} teach a redundant robot how to perform various
%object manipulation tasks by adopting a co-active feedback-based
%learning framework.
However, due to the presence of at least two
agents in these scenarios and the complex embodiment of the robot,
these are difficult tasks characterized by a very large
state-space. Additionally, during the execution of the policy,
guarantees on the correct interpretation of spatial constraints that
arise from social rules may be required to make the robot acceptable
in human populated environments.  Although policy learning methods
have shown promising results in robotics~\cite{Bagnell2013}, their
application to problems with large state spaces is often impractical
and computationally expensive. In fact, in order to compute a good
policy, the state space needs to be exhaustively explored and the
curse-of-dimensionality problem typically occurs. To mitigate this
issue, a frequent solution adopted in literature is that of
initializing the policy to imitate a sub-optimal expert and
consequently avoid the initial (expensive) exploration phase. For
example, Kober and Peters~\cite{Kober2009}, learn a ball-in-a-cup task
by first initializing motion primitives through imitation, and then
improving robot policies via episodic learning. Similarly, Argall et
al.~\cite{Argall2010} take advantage of user tactile feedbacks to
influence the learning process and refine a demonstrated
policy. In~\cite{Riccio2016a}, instead, the authors learn the policy
of a robot soccer defender by imitating the strategy of the opponent
players and incrementally improving it through several Monte-Carlo
rollouts. However, in more complex scenarios, (1) obtaining a
sufficient number of expert demonstrations is time consuming or
impossible, (2) depending on the embodiment of the robot, such
demonstrations cannot be easily mapped into the robot configuration --
e.g., in the case of highly redundant platforms.

While we use data aggregation~\cite{Ross2011,Ross2014} to
incrementally learn a policy, in our problem we do not initialize the
robot behavior by imitation due to the interactive nature of handover
tasks and to the unpredictability of human behaviors. Conversely, we
allow our robot to learn from experience and to improve its behavior
while operating. To avoid the curse-of-dimensionality, we restrict the
search space by leveraging the concept of spatio-temporal affordances,
that encode the semantics of robot actions. Affordances have been
originally defined as action \emph{opportunities} that objects offer
to agents~\cite{Gibson1979} and, in robotics, they have been adopted
to better represent objects and their related actions. This concept
has also been recently extended to describe environments as a
combination of spatial affordances to adapt robot behaviors. For
example, in~\cite{Epstein2015} and~\cite{Luber2011}, the authors use
spatial affordances to support robot movements in a navigation task
and to improve the performance of a tracking system respectively. In
Kapadia et al.~\cite{Kapadia2009} affordances are used to select the
best action for collision avoidance, while in Diego et
al.~\cite{Diego2011} they are used to navigate in crowded areas. Since
the above cited works only represent spatial affordances to improve
robot navigation skills, their approach cannot be generalized to
encode spatial semantics of different tasks. Instead, we leverage the
formalization of spatio-temporal affordances proposed
in~\cite{Riccio2016} and we continuously improve the robot policy by
exploiting the constraints imposed by spatio-temporal affordances.

The idea of improving robot policies based on object
affordances~\cite{Lopes2007,Katz2013}, or discovering object
affordances given an initial policy~\cite{Wang2014} has been already
investigated in recent works. For example, in~\cite{Wang2014} the
authors exploit a simple policy to learn the affordance of an object
to be pushed. In~\cite{Lopes2007} and~\cite{Katz2013}, instead,
affordances have been respectively exploited to learn action
primitives for the imitation of humans and for autonomous pile
manipulation. Likewise, we combine the use of affordances and policy
improvement methods. However, while refining the robot behavior, we
also modify the affordance model, depending on the outcomes of the
current policy. In this way, on the one hand a semantic representation
of the space in the environment is incrementally learned, on the other
hand such representation is used to reduce the search space for the
policy improvement.

Summarizing, the contribution of this paper is twofold: (1) we use
spatial affordances to efficiently reduce the search space for
learning the policy of a complex task; (2) our algorithm
simultaneously improves the robot policy and refines the spatial
affordance model.

%%%%%%%%%%%%%%%%%%%%%%%%%%%%%%%%%%%%%%%%%%%%%%%%%%%%%%%%%%%%%%%%%%%%%%%%%%%%%%%%

\section{Spatio-Temporal Affordances}
\label{sec:stam}

In this paper we consider an extension of the spatial affordance
theory based on the concepts of spatial semantics. More precisely, we
establish a connection between the environment (with its operational
functionality) and spatio-temporal affordances. According to the
definition provided in~\cite{Riccio2016} and slightly modified in
Def.~\ref{def:sta}, a Spatio-Temporal Affordance (STA) is a function
that defines areas of the operational environment that afford an
action, given a particular state of the world.

\begin{defn}
  A \textit{Spatio-Temporal Affordance (STA)} is a function
  \begin{align}
    f_{E, \set{T}}: S \times \Theta \rightarrow \mathrm{A}_E.
  \end{align}
  $f_{E, \set{T}}$ depends on the environment $E$ and a set of actions
  or tasks $\set{T} = \lbrace\tau(t)\rbrace$ to be performed. It takes
  as input the state of the environment $s_E(t) \in S$ at time $t$, a
  set of parameters $ \boldsymbol{\theta} \in \Theta$ characterizing
  the affordance function, and outputs a map of the environment
  $\mathrm{A}_E$ that evaluates the likelihood of each area of $E$ to
  afford $\set{T}$ in $s_E$ at time $t$.
  \label{def:sta}
\end{defn}

The function $f_{E, \set{T}}$ characterizes spatial semantics by
evaluating areas of $E$ where the tasks $\set{T}$ can be afforded and
generating, at each time $t$, the affordance spatial distribution as a
map $\mathrm{A}_E$. Then, a representation of the STA of an
environment can be learned, updated and used by an autonomous agent
through a Spatio-Temporal Affordance Map (STAM). 
The core element of a STAM is the function $f_{E,
  \set{T}}$, that takes as input the parameters $\boldsymbol{\theta}$
obtained from an \textit{affordance description module}, as well as
the current state of the world from the \textit{environment module},
which also provides the set of tasks $\set{T}$ to be considered.
In particular, the affordance description module is a
  knowledge base composed by a library of parameters
  $\boldsymbol{\theta}$ that represent the \textit{signature} of the
  STA. The signature defines the spatial distribution of affordances
  within the environment.
  The \textit{environment module}, instead, encodes the
  state of the world $s_E(t)$ and provides such a state to the STA
  function, together with the set of tasks $\set{T}$ to be considered.

\begin{figure}[t!]
  \centering
  \includegraphics[width=0.9\columnwidth, trim=3.9cm 2.4cm 2cm 4cm, clip]{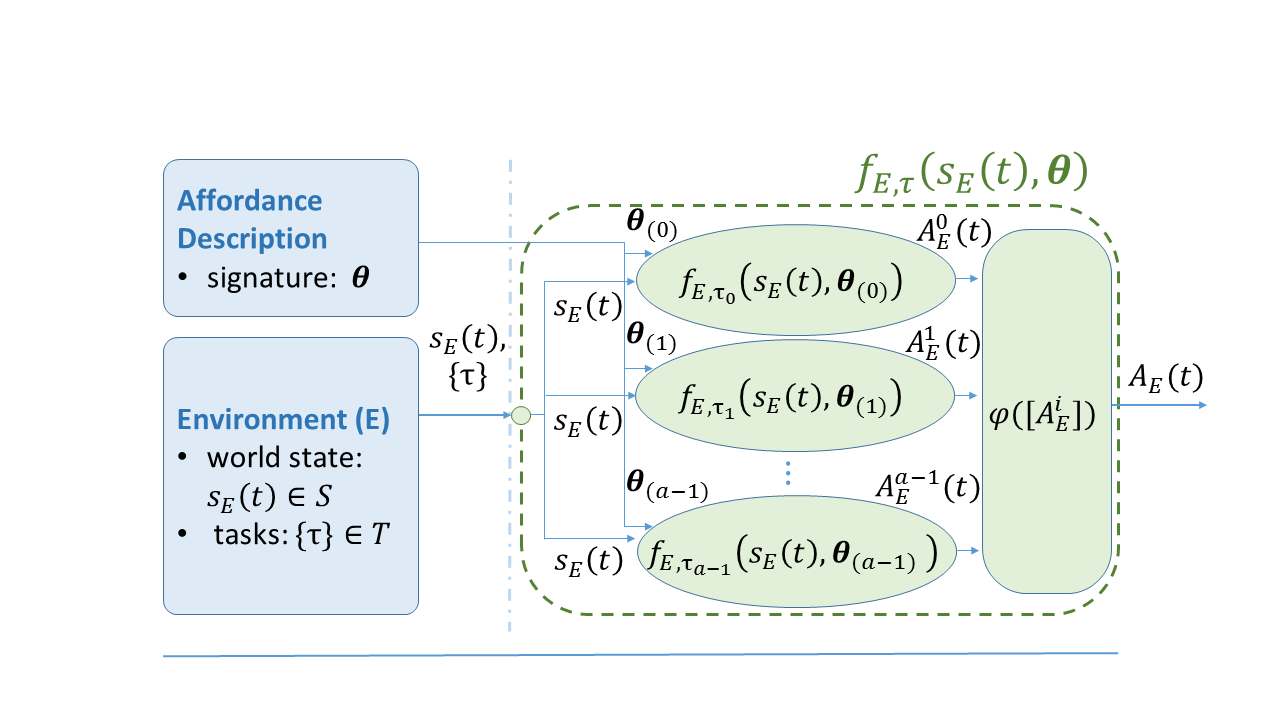}
  \caption{Decomposition and detailed overview of a Spatio-Temporal
    Affordance Map (STAM).}
  \label{fig:sta_multi}
\end{figure}
%\begin{itemize}
%\item the affordance description module (\textbf{a-module}) is a
%  knowledge base composed by a library of parameters
%  $\boldsymbol{\theta}$ that represent the \textit{signature} of the
%  STA. The signature defines the spatial distribution of affordances
%  within the environment;
%
%\item the \textit{environment module} (\textbf{e-module}) encodes the
%  state of the world $s_E(t)$ and provides such a state to the STA
%  function, together with the set of tasks $\set{T}$ to be considered.
%\end{itemize} 
STAM can be used both to interpret relations among different
affordances (if they exist) and represent affordances individually. In
fact, as shown in \figurename~\ref{fig:sta_multi}, a STA can be seen
as a composition of different $f_{E, \tau_j}(s_{E}(t),
\boldsymbol{\theta}(j))$ functions, with $j \in [0, \alpha-1]$, where
$\alpha$ is the number of affordances, each modeling the spatial
distribution $\mathrm{A}_E^j$ of a particular affordance in $E$. These
are then combined by a function $\phi$, that takes as input all the
$\mathrm{A}_E^j$ and outputs a map $\mathrm{A}_E$ that satisfies
$\set{T}$, according to the considered affordances. 

In the case of a humanoid robot performing handovers, the STAM
architecture enables the agent to formalize the problem with a
different STA for each action. Let us assume, in this example, to have
only two possible actions: ``go forward'' and ``left arm
forward''. The two affordance functions encode the spatial
distributions of supporting each action independently from each
other. For example, the ``go forward'' can be executed when the robot
is far from the target, while the ``left arm forward'' can be
performed when they are close enough. Then, in order to respect all
the constraints and perform the handover, STAM generates a unique map
$\mathrm{A}_E$ by means of $\phi$. In our case, such function reflects
the support of the affordance for the handover task as a combination
of the two actions~\cite{Riccio2016}.

%%%%%%%%%%%%%%%%%%%%%%%%%%%%%%%%%%%%%%%%%%%%%%%%%%%%%%%%%%%%%%%%%%%%%%%%%%%%%%%%

\section{$\pi$-STAM}
\label{sec:pi_stam}

\begin{figure}[t!]
  \centering
  \includegraphics[width=0.6\columnwidth]{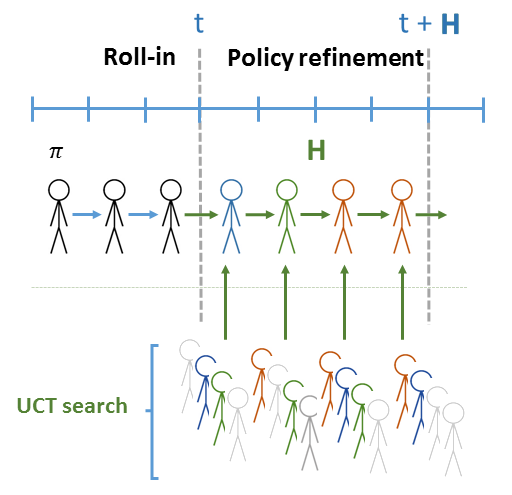}
  \caption{UCT execution at a state $s_t$ generated by following the
    policy $\pi_{i-1}$ (roll-in). The state is expanded by the
    algorithm for $H$ iterations. In the figure, different colors of
    the agents represent different actions. In particular, gray
    actions are considered to be non-legal in $s_{t+h-1}$ and, hence,
    they are not evaluated. UCT, then, expands the best action
    according to Eq.\ref{eq:1}.}
  \label{fig:uct_alg}
\end{figure}

We formalize our learning problem by adopting the Markov Decision
Process notation, where $S$ represents the continuous state space of
the environment, $A$ consists of a discrete set of robot actions, $T:
S \times A \rightarrow S$ is the transition function and $R(s)$ is the
immediate reward obtained in state $s \in S$, bounded in $[0, 1]$. Due
to its complexity, we access the dynamics of the world only through
samples obtained by executing a policy in the world or in a simulator.

The goal of our algorithm, Policy Improvement with Spatio-Temporal
Affordances ($\pi$-STAM), is to generate, at each iteration $i \in
\{0, ..., N\}$, a new policy $\pi_i$ that improves $\pi_{i-1}$, with
$\pi_0$ initialized to be random. More in detail, as described in
Algorithm~\ref{alg:pistam}, $\pi$-STAM takes as input a set $\set{D}$
of random state-action pairs to generate an initial policy $\pi_0$ and
affordance signature $\boldsymbol{\theta}_0$ of the STA function
$f^{0}_{E, \set{A}}$. As in previous
work~\cite{Ross2011,Ross2014,Chang2015,Riccio2016a}, the dataset is
used to generate a policy $\pi$ through a supervised learning approach
-- more specifically a (generic) classifier. Then, during each
iteration, the algorithm proceeds as follows:

\begin{enumerate}
[leftmargin=0cm,itemindent=.5cm,labelwidth=\itemindent,labelsep=0cm,align=left]

\item it executes the previous policy $\pi_{i-1}$ for $T$ timesteps
  and generates a set of $T$ states $\{s_t \mid t = 1 \dots T\}$;
\item it runs, for each $s_t$, the Upper Confidence Bound for Trees
  (UCT) algorithm~\cite{Browne2012} -- a variant of Monte Carlo Tree
  Search. In particular, as shown in \figurename~\ref{fig:uct_alg},
  UCT is an iterative algorithm that, at each iteration $h = 1 \dots
  H$, simulates the execution of each \textit{legal action} in
  $s_{t+(h-1)}$ and selects the best action $a^*_h$ as
  \begin{align}
    \label{eq:1}
    e &= C \cdot \sqrt{\frac{\log(\sum_a{n(s_{t+(h-1)},a)})}{n(s_{t+(h-1)},a)}}\\
    a^*_h &= \argmax_a V_i(T(s_{t+(h-1)}, a)) + e,
  \end{align}
  where $V_i(T(s_{t+(h-1)}, a))$ is the value function of the state
  $s' = T(s_{t+(h-1)}, a)$ obtained by executing action $a$ in state
  $s_{t+(h-1)}$, $C$ is a constant that multiplies and controls the
  exploration term $e$, and $n(s_{t+(h-1)},a)$ is the number of
  occurrences of $a$ in $s_{t+(h-1)}$. The value function $V_i(s)$ is
  obtained by back-propagating the final reward of each simulation to
  all the traversed states $s$. To compute occurrences, since the
  state space $\set{S}$ is continuous, we define a comparison operator
  that returns true whenever the difference between two states is
  smaller than a threshold $\rho$. After selecting the best action, a
  new state is generated as $s_{t+h} = T(s_{t+(h-1)}, a^*_h)$ and all
  the obtained state-action pairs $\{(s_{t+(h-1)}, a^*_h) \mid h = 1
  \dots H\}$ are collected in a dataset $\set{D}_{new}$. Legal actions
  in $s_{t+(h-1)}$ are selected from $\set{A}$ according to the STA
  function $f^{i-1}_{E, \set{A}}((s_{t+(h-1)},
  \boldsymbol{\theta}_{i-1})$. In particular, actions are considered
  to be legal if their likelihood to be afforded in $s_{t+(h-1)}$ is
  higher than a threshold $\lambda$ that is adaptively determined
  according to
  \begin{align}
    \label{eq:2}
    \lambda &= \frac{1}{2}f^{i-1}_{max} = \frac{1}{2}\max\limits_{a
      \in \set{A}}f^{i-1}_{E, a}(s_{t+(h-1)}, \boldsymbol{\theta}_{i-1}),
  \end{align}
 
  where $f^{i-1}_{max}$ is the maximum affordance function value with
  respect to the actions $a \in \set{A}$. Additionally, actions whose
  likelihood is lower than $\lambda$ can be randomly selected to be
  legal with $\epsilon$ probability. In fact, this is useful to avoid
  overfitting to wrong affordance models especially during the first
  iterations of $\pi$-STAM.

\item it aggregates the new state-action pairs contained in
  $\set{D}_{new}$ to the previous dataset $\set{D}$. Note that, (1)
  the dataset $\set{D}$ is non-i.i.d., since chosen actions influence
  the distribution of states, and (2) while aggregating new
  state-action pairs, previously seen states are removed from the
  original dataset to avoid duplicates with different action
  labels. Also in this case, since the state space $\set{S}$ is
  continuous, we use the comparison operator previously introduced;
\item it uses the aggregated dataset $\set{D}$ to train a new
  classifier $\pi_i$ that substitutes the policy used at the previous
  iteration, as well as the new signature $\boldsymbol{\theta}_i$ of
  the STA function $f^i_{E, \set{A}}$. As shown in Eq.~\ref{eq:3},
  such signature is chosen to maximize, for each action $a \in
  \set{A}$, the summed likelihoods to afford $a$ in all the states
  labeled with $a$ in $\set{D}$:
  \begin{align}
    \label{eq:3}
    \boldsymbol{\theta}(j)_i = \argmax_{\boldsymbol{\theta}(j)}
    \sum_{\{s \mid (s,a_j) \in
      \set{D}\}} &f_{E,a_j}(s_t,\boldsymbol{\theta}(j)),
  \end{align}
  where $j = 1 \dots |\set{A}|$. This optimization consists of finding
  our parameters $\boldsymbol{\theta}(j)_i$ by maximizing the
  likelihood of the portion of the dataset labeled with the considered
  action $a_j$.
\end{enumerate}

Note that, in this work, we assume to have an affordance for each
action and the set of tasks $\set{T}$ corresponds to the set of
actions $\set{A}$.

\begin{algorithm}[t]
  \DontPrintSemicolon \SetAlgoLined 
  \SetNlSty{}{}{)}
  \SetAlgoNlRelativeSize{0}

  \KwIn{$\set{D}_0$: dataset of random state action pairs $\{s, a\}$}

  \KwOut{$\pi_N$: policy learned after N iterations of the algorithm,
    $\boldsymbol{\theta}_N$: signature of the STA function learned
    after N iterations of $\pi$-STAM.}

  \KwData{$\set{A}$: discrete set of robot actions, $N$: number of
    iterations of the algorithm, $\Delta$: initial state distribution,
    $H$: UCT horizon, $T$: policy execution timesteps.}
  
  \Begin{
    Train classifier $\pi_0$ on $\set{D}_0$.\;
    \For{$j = 1 \dots |\set{A}|$} {
      Initialize $\boldsymbol{\theta}(j)_0 \leftarrow
      \argmax_{\boldsymbol{\theta}(j)} \sum_{\{s \mid (s,a_j) \in
        \set{D}_0\}} f_{E,a_j}(s_t,\boldsymbol{\theta}(j))$.\;
    }
    Initialize $\set{D} \leftarrow \set{D}_0$.\;
    \For{$i = 1$ \KwTo $N$}{
      $s_0 \leftarrow$ random state from $\Delta$.\;
      \For{$t = 1$ \KwTo $T$}{
        \nl Get state $s_t$ by executing $\tilde{\pi}_{i-1}(s_{t-1})$.\;
        \nl $\set{D}_{new} \leftarrow$ UCT($\set{A}$, $s_t$).\;
        \nl $\set{D} \leftarrow \set{D} \cup \set{D}_{new}$.\;
      }

      \nl Train classifier $\pi_{i}$ on $\set{D}$.\;
      \For{$j = 1 \dots |\set{A}|$} {
        $\boldsymbol{\theta}(j)_i \leftarrow
        \argmax_{\boldsymbol{\theta}(j)} \sum_{\{s \mid (s,a_j) \in
          \set{D}\}} f_{E,a_j}(s_t,\boldsymbol{\theta}(j))$.\;
      }
    }
    \BlankLine
    \Return{$\pi_{N}$, $\boldsymbol{\theta}_N$}\;
  }
  \caption{$\pi$-STAM}
  \label{alg:pistam}
\end{algorithm}

To enforce prior knowledge, the signature of the affordance function
can be initialized by providing to $\pi$-STAM a dataset $\set{D}_0$
representing the desired action semantics. This can be simply achieved
by collecting example demonstrations from a human, or by manually
constructing $\set{D}_0$ - when possible - with few salient examples.

%%%%%%%%%%%%%%%%%%%%%%%%%%%%%%%%%%%%%%%%%%%%%%%%%%%%%%%%%%%%%%%%%%%%%%%%%%%%%%%%

\section{Experimental Evaluation}
\label{sec:evaluation}

Human-robot handovers are difficult tasks, where the large state space
of the problem greatly affects the usability of policy learning
methods. Since the goal of $\pi$-STAM is to refine an affordance model
that reduces the search space of the policy, in this section we
evaluate our method after a small number iterations of the
algorithm. In particular, we test the effectiveness of $\pi$-STAM to
obtain a good policy and yet to reduce (1) the number of simulated
action executions in UCT and (2) the duration of an iteration of the
algorithm. To this end, we compare against a baseline where
affordances are not considered and all the actions are always
legal. We evaluate the learned policies both on a simulated and real
NAO robot by using the average reward obtained over 10 different
trials. Additionally, we qualitatively evaluate the learned affordance
models by observing their evolution as a function of
%both the iterations of $\pi$-STAM and 
the state of the world. Finally, we show an example of prior knowledge
introduced in the affordance model. In particular, we embed in the
model the ``eye-contact'' social rule, according to which no handover
can be performed when the partner is not paying attention.

%%%%%%%%%%%%%%%%%%%%%%%%%%%%%%%%%%%%%%%%%%%%%%%%%%%%%%%%%%%%%%%%%%%%%%%%%%%

\subsection{Experimental Setup}
\label{subsec:setup}

In our experimental setup, the task of the humanoid robot is to take
an object -- a red ball -- from the hands of a human
operator. Experiments have been executed both on a real NAO platform
(V4) and on the V-REP simulator running on a single Intel Core
i7-5700HQ core, with CPU@2.70GHz and 16GB of RAM. During the learning
phase of the algorithm (1) the horizon is selected to be $H = 4$ to
guarantee a good trade-off between performance improvement and
usability of the approach, (2) the probability to randomly expand a
non-legal action in UCT is set to $\epsilon = 0.3$, (3) the affordance
function is implemented by Gaussian Mixture Models (GMMs) and (4) the
policy is learned through GMM classification. The signature
$\boldsymbol{\theta}$ of the STA function is hence composed as a tuple
$\boldsymbol{\theta} = \langle \pi_1, \mu_1, \Sigma_1, \dots, \pi_N,
\mu_N, \Sigma_N \rangle$, where $\pi_i$ is the prior, $\mu_i$ the mean
vector and $\Sigma_i$ the covariance matrix of a mixture of $N$
Gaussians. The state of the problem is composed of the Cartesian pose
(position and angles) of the robot kinematic chains corresponding to
the head and the two arms, together with the state of the hands
(opened/closed). Additionally, the state includes the relative
distance between the robot and object poses, the position of the
target in the image frame of the cameras as well as an ``attention
bit'' indicating if the human is looking towards the NAO. The robot is
allowed to execute the following 27 actions: (1) rotate the head to
the left, right, up and down; (2) move the body forward, backward,
left, right and rotate it on the left and right; (3) move one of the
two arms forward, backward, left, right, up and down; (4) close and
open the hands; (5) execute the ``null'' action to stay still. The
reward function $R(s)$, instead, is modeled to penalize the distance
of the robot from the target, as well as orientations of the head
where the object is not centered in one of the two
cameras. Additionally, the reward penalizes the robot whenever its
hands are closed before the object is reached.

%%%%%%%%%%%%%%%%%%%%%%%%%%%%%%%%%%%%%%%%%%%%%%%%%%%%%%%%%%%%%%%%%%%%%%%%%%%

\subsection{Results}
\label{subsec:results}

\begin{figure}[t!]
  \centering
  \subfigure[Simulation]{
    \includegraphics[width=0.9\columnwidth]{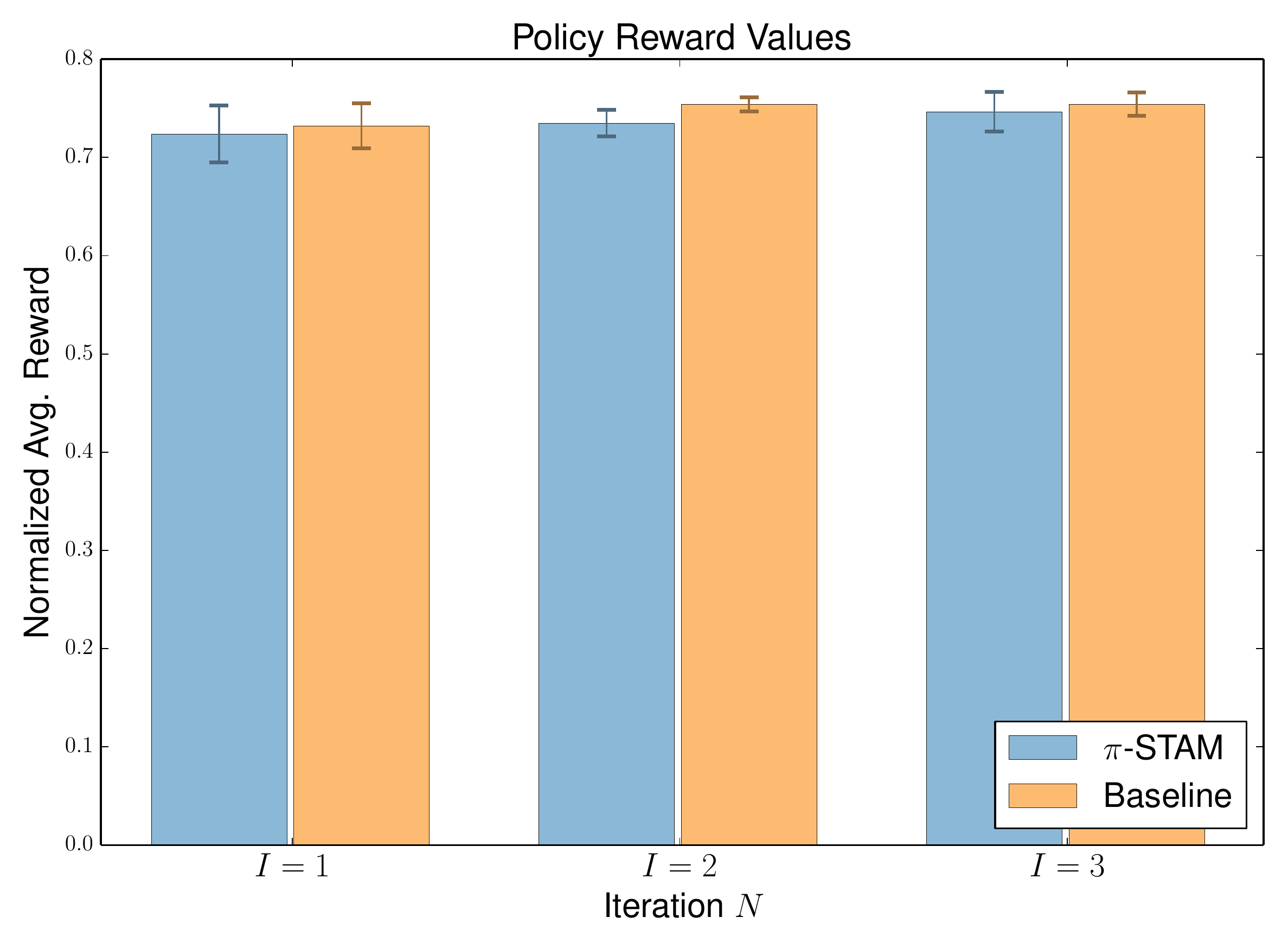}
    \label{fig:sim_rewards}
  }
  \subfigure[Real-robot]{
    \includegraphics[width=0.9\columnwidth]{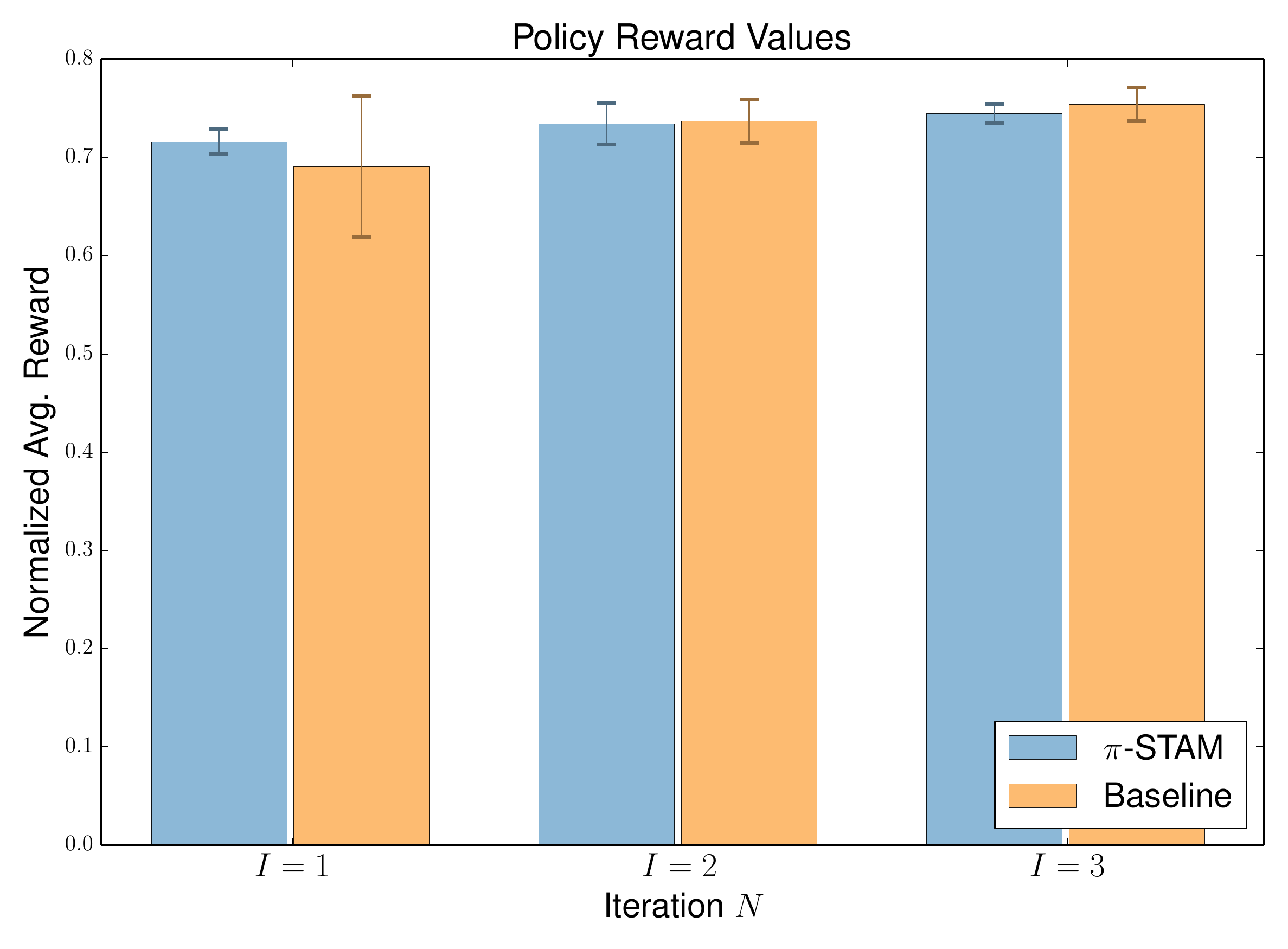}
    \label{fig:real_rewards}
  }
  \caption{Normalized average reward obtained by $\pi$-STAM and the
    baseline algorithm over 10 simulated handovers, both in simulation
    and real-robot scenario.}
  \label{fig:rewards}
\end{figure}
\begin{figure}[t!]
  \centering
  \subfigure[Affordance and random expanded states]{
    \includegraphics[width=0.9\columnwidth]{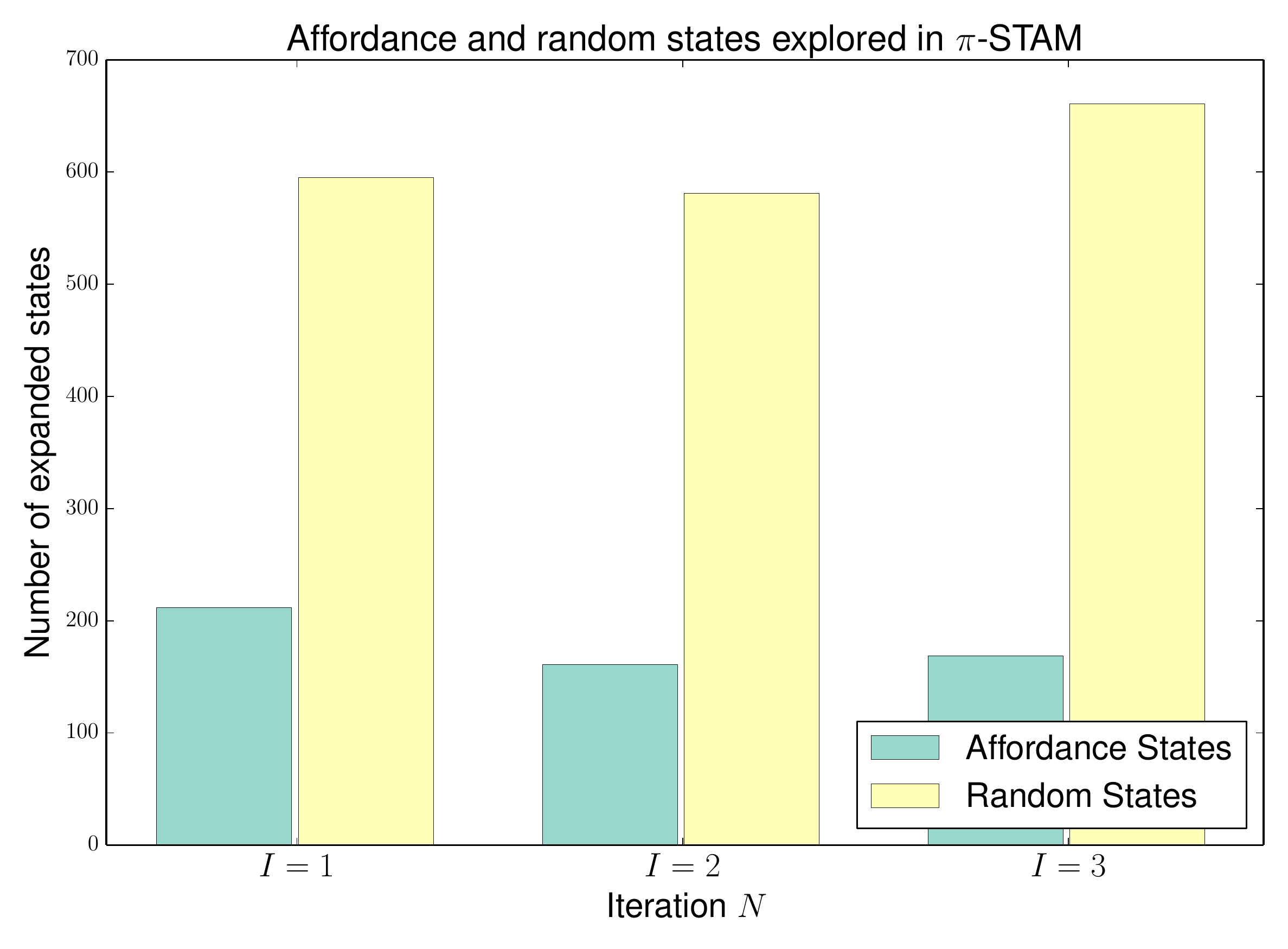}
    \label{fig:affordance_random_states}
  }
  \subfigure[Total expanded states]{
    \includegraphics[width=0.9\columnwidth]{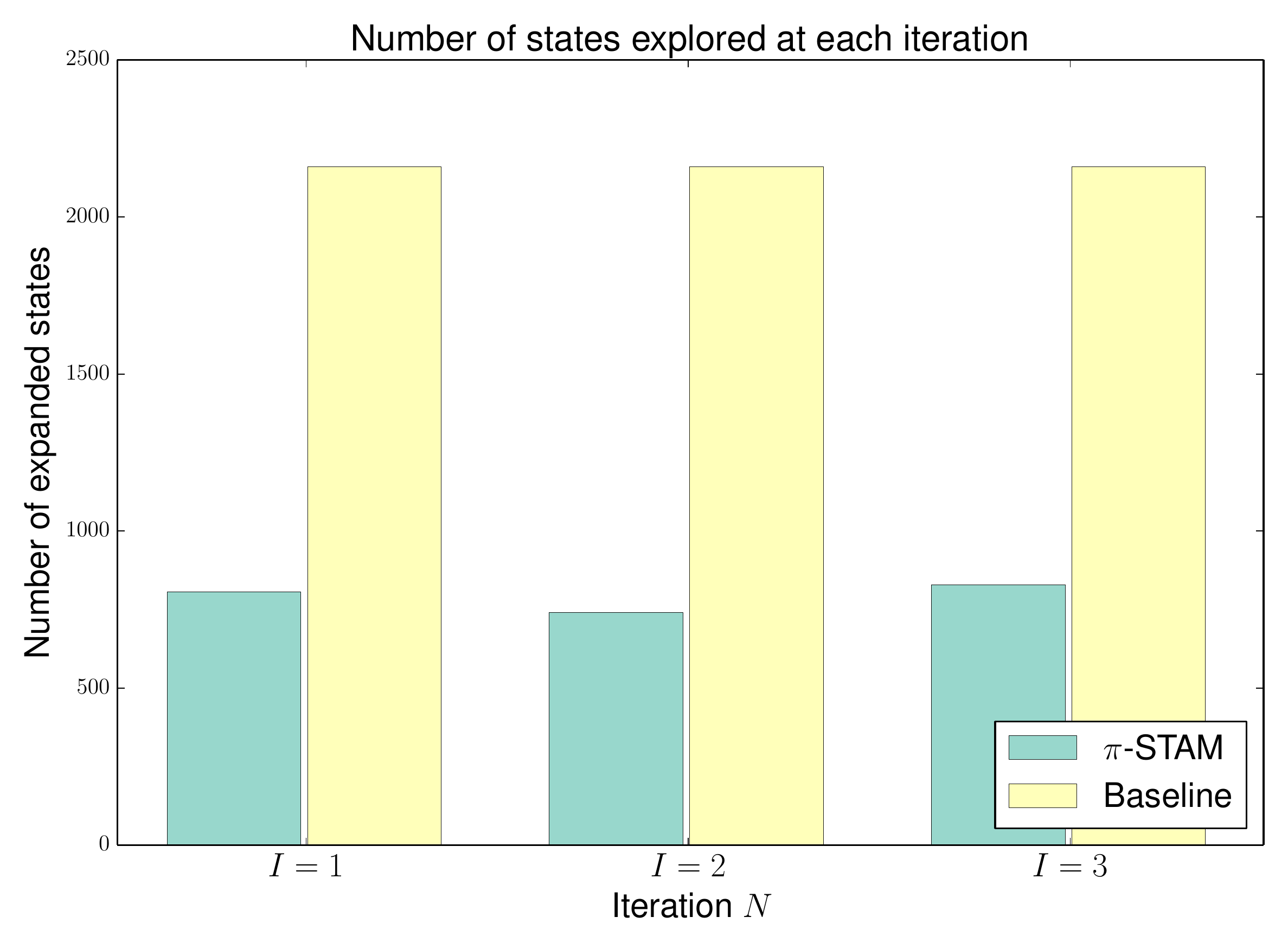}
    \label{fig:total_states}
  }
  \caption{Comparison between the mean number of states expanded by
    UTC in $\pi$-STAM and in the baseline algorithm. As shown in the
    top figure, in the former algorithm states are expanded as legal
    actions because of affordances or because they are randomly
    selected. In the latter case, instead, the reported value is
    constant because all actions are always evaluated.}
  \label{fig:visited_states}
\end{figure}
\begin{figure}[t!]
  \centering
  \includegraphics[width=0.9\columnwidth]{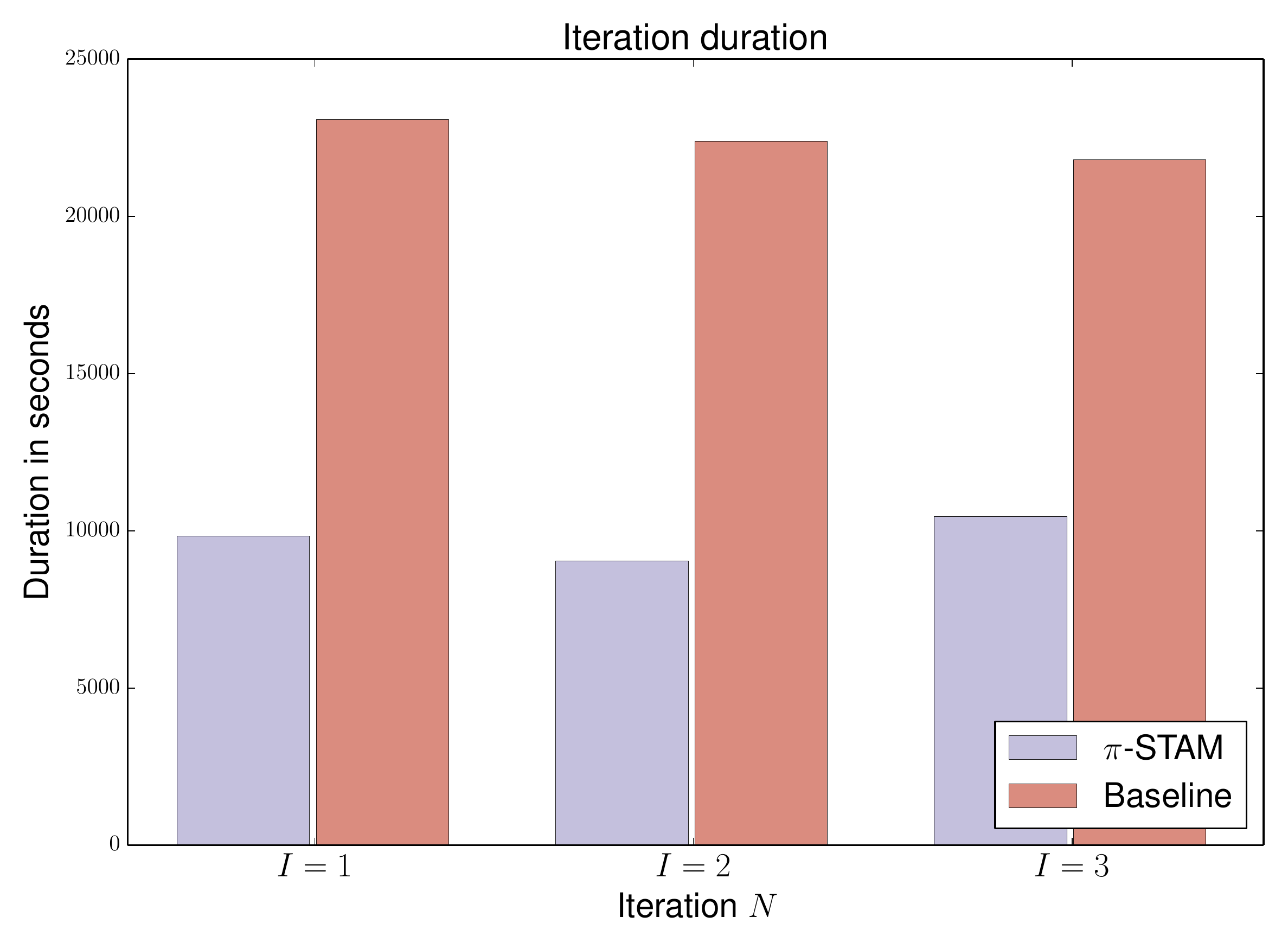}
  \caption{Comparison between the time consumption (in seconds) of
    3 iterations of $\pi$-STAM and the baseline algorithm.}
  \label{fig:time_consumptions}
\end{figure}

The goal of our robot is to perform handovers and maximize the
obtained reward, that is a measure of ``goodness'' of the executed
policy. In our experiments, we analyze the average reward obtained by
the agent, as well as its standard deviation, both in simulation and
on a real-robot scenario. In particular, \figurename~\ref{fig:rewards}
shows on the y-axis the normalized average reward obtained by a NAO
during 10 handover trials both in simulation
(\figurename~\ref{fig:sim_rewards}) and on the real robot
(\figurename~\ref{fig:real_rewards}). In the figure, we compare the
reward obtained by $\pi$-STAM and the baseline algorithm -- that does
not use affordances and always considers all actions to be legal --
over 3 iterations of the algorithms. It is worth to notice that, as
shown in \figurename~\ref{fig:visited_states}, while the baseline
fully explores all the actions during the execution of UCT, $\pi$-STAM
only expands legal actions. Still, the difference in the obtained
average reward only slightly favors the baseline algorithm and the
generated policies perform similarly. Conversely, the computational
load (\figurename~\ref{fig:visited_states}) and time consumption
(\figurename~\ref{fig:time_consumptions}) of $\pi$-STAM are
significantly reduced with respect to the baseline. In particular, the
number of states that $\pi$-STAM expands because of affordances
(\figurename~\ref{fig:affordance_random_states}) is small. This
demonstrates that $\pi$-STAM is able to efficiently capture the
semantics of actions and, thanks to this, the policy search algorithm
only spends time in ``good'' portions of the state space -- those with
higher expected reward. Notice that, as shown in
\figurename~\ref{fig:affordance_random_states}, a significant part of
the states expanded during the execution of $\pi$-STAM is randomly
selected and depends on the chosen $\epsilon$ value. While this
represents a lower-bound to the number of expanded states, it is
essential to guarantee exploration in the affordance model and avoid
overfitting to wrong state spaces during the first iterations of the
algorithm. Although not discussed here in detail, it is possible to
improve the efficiency of the algorithm by reducing the $\epsilon$
value when the number of iterations of $\pi$-STAM
increases. Intuitively, this enables the algorithm to first explore
the state space and then to follow the learned affordance model
proportionally to a confidence that increases over the number of
iterations.

\begin{figure}[t!]
  \centering
  \subfigure[Body forward]{
    \includegraphics[width=0.25\columnwidth]{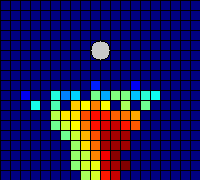}
    \label{fig:aff_body_forward}
  }
  \subfigure[Head right]{
    \includegraphics[width=0.25\columnwidth]{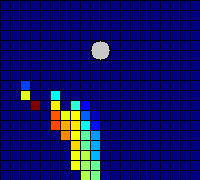}
    \label{fig:aff_head_right}
  }
  \subfigure[Head up]{
    \includegraphics[width=0.25\columnwidth]{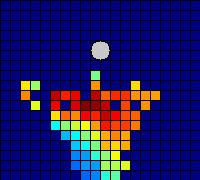}
    \label{fig:aff_head_up}
  }
  \caption{Heat-map of the spatial affordance distribution generated
    by $\pi$-STAM (after 3 iterations) for the body forward, head
    right and head up actions. The initial distribution was composed
    of states located at a distance between 45cm and 60cm in front of
    the target. Each cell of the grid has a size of approximately
    5cm. Red colors represent high affordance values in the cell,
    while blue colors represent low values.}
  \label{fig:aff_distributions}
\end{figure}

\begin{figure}[t!]
  \centering
  \subfigure[No eye-contact]{
    \includegraphics[width=0.85\columnwidth]{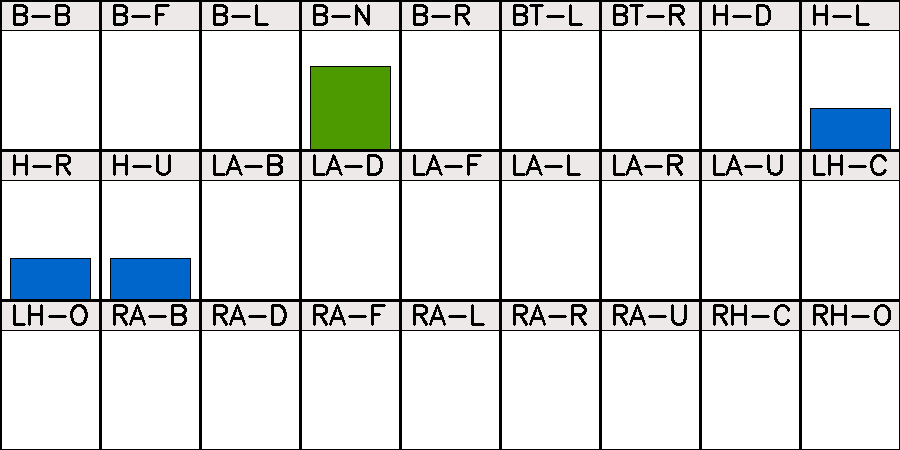}
    \label{fig:no_attention}
  }
  \subfigure[Eye-contact]{
    \includegraphics[width=0.85\columnwidth]{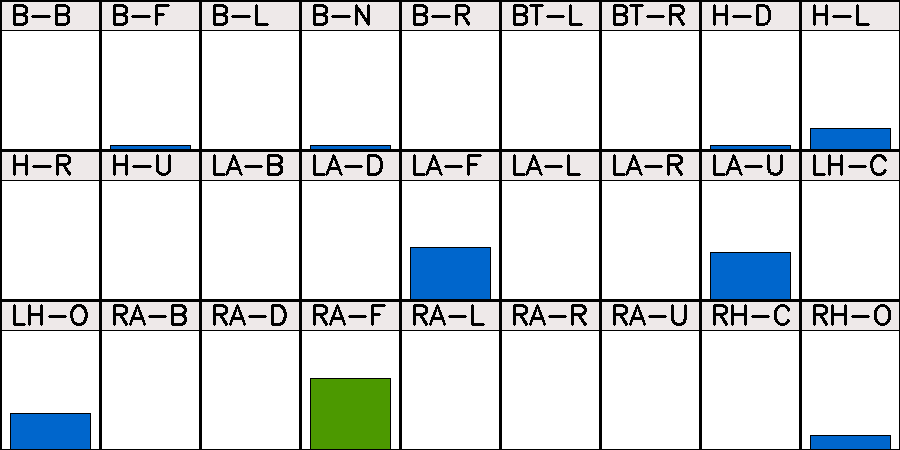}
    \label{fig:attention}
  }
  \caption{Affordance values for all the actions when the
    ``eye contact'' social rule is not respected and viceversa. The
    green bar corresponds to the maximum affordance action. Each
    action is named with the first letter corresponding to the
    considered kinematic chain (e.g. RA: right arm), while the second
    letter corresponds to the movement direction (e.g., B: backward,
    U: up, etc.). B-N, instead, corresponds to the ``null'' action.}
  \label{fig:social_rule}
\end{figure}

In addition to the learned policy, we qualitatively evaluate the
obtained affordance model. \figurename~\ref{fig:aff_distributions},
for example, shows the heat-map -- with a 5cm granularity --
corresponding to the spatial affordance distribution of three actions:
body forward (\figurename~\ref{fig:aff_body_forward}), head right
(\figurename~\ref{fig:aff_head_right}) and head left
(\figurename~\ref{fig:aff_head_up}). These have been generated from
the model learned after 3 iterations of $\pi$-STAM with a small
initial distribution composed of states located at a distance between
45cm and 60cm in front of a fixed target. From the figures it can be
observed that the learned affordances accurately model the need of the
robot to (1) go forward when it is far from the target and stop at a
distance of $~$20-25cm (from which the target is reachable with the
arms), (2) turn the head right when it reaches positions on the left
of the target, (3) increase the head pitch when it is close to the
target (located at a position higher than the robot). Finally,
\figurename~\ref{fig:social_rule} shows the affordances of each action
when implementing an ``eye contact'' social rule. Such rule states
that the robot should wait for eye contact with the human partner to
start the handover. In this test, we simply assume to have eye contact
when the (Aldebaran) tracker of the NAO detects a face that is
oriented towards the robot. As shown by results in
\figurename~\ref{fig:no_attention}, when the human is not paying
attention, the highest affordance value (in green) corresponds to the
``null'' action. Additionally, head movements are allowed to search
for eye contact. Conversely, when the partner looks at the NAO, the
``right arm forward'' action has the highest affordance value (see
\figurename~\ref{fig:attention}), and other arm-related actions are
enabled. To provide this prior knowledge to the system, we initialize
the signature $\boldsymbol{\theta}_0$ of the STA function with
parameters of the GMM learned from a dataset where (1) all the actions
are enabled when the ``attention bit'' is on and (2) only the head
rotations and ``null'' actions are allowed when the ``attention bit''
is off.

%%%%%%%%%%%%%%%%%%%%%%%%%%%%%%%%%%%%%%%%%%%%%%%%%%%%%%%%%%%%%%%%%%%%%%%%%%%%%%%%

\section{Discussion}
\label{sec:discussion}

In this paper we presented $\pi$-STAM, an algorithm that
simultaneously improves the robot policy and refines the spatial
affordance function to model action semantics. We used and evaluated
our method on the execution of human-robot handovers. Our experiments
show that, by using spatial affordances to efficiently reduce the
search space, $\pi$-STAM is able to rapidly generate a good policy for
a complex task.
\paragraph*{Contributions}
%\subsection*{Contributions}
%\label{sec:contributions}
The main contribution of this paper consists in combining affordances
and policy improvement methods in such a way that, while the robot
behavior is refined, also the affordance model is modified depending
on the outcomes of the current policy. The combination of UCT and
affordances results in a practical algorithm, that allows a real-world
implementation on complex robot domains. By relying on Spatio-Temporal
Affordance Maps, social rules can be enforced in the policy
generation. Additionally, by using data aggregation, $\pi$-STAM
preserves the main characteristics of algorithms like
\textsc{AggreVate} and can be easily transformed to an online method.
\paragraph*{Limitations and Future Work}
%\subsection*{Limitations and Future Work}
%\label{sec:limitations}
Our algorithm still presents some limitations, due to the use of
expensive calls to a simulator. For this reason, even if UCT and
affordances make $\pi$-STAM practical, the algorithm is less appealing
when the available computational resources are very limited. For this
reason, in our future work we plan to reduce the use of a simulator by
combining online and offline knowledge in
UCT~\cite{Gelly2007}. Additionally, we plan to use affordances to
establish a hierarchy and priority among actions.

%%%%%%%%%%%%%%%%%%%%%%%%%%%%%%%%%%%%%%%%%%%%%%%%%%%%%%%%%%%%%%%%%%%%%%%%%%%%%%%%

\bibliographystyle{IEEEtran}
\bibliography{references}

\balance
\end{document}